\documentclass[letterpaper, 10 pt, conference]{ieeeconf}
\IEEEoverridecommandlockouts    %
\usepackage[T1]{fontenc}
\usepackage{newtxtext,newtxmath}
\usepackage{microtype}

\usepackage{graphics}
\usepackage{graphicx}
\usepackage{amsmath}
\usepackage{booktabs}
\usepackage{array}          %
\usepackage{color}
\usepackage{hyperref}
\usepackage{cite}
\usepackage{xspace}
\usepackage{placeins}
\DeclareMathOperator*{\argmax}{arg\,max}

\title{\LARGE \bf Selective Sensor Acquisition for Multimodal Place Recognition}

\author{Margaux Edwards, Tobias Fischer, and Michael Milford%
\thanks{The authors are with the QUT Centre for Robotics, School of Electrical Engineering and Robotics, Queensland University of Technology, Brisbane, QLD 4000, Australia. Email: margaux.edwards@hdr.qut.edu.au}
}

\begin{document}
\microtypesetup{activate=false}
\raggedbottom
\thispagestyle{empty}
\pagestyle{empty}

\maketitle

\microtypesetup{activate=true}
\begin{abstract}
Multimodal Place Recognition (PR) can improve robustness, but most systems rigidly read and process every sensor for every query, even when the primary sensor is sufficient, with implications for energy consumption, second sensor update rate, and domain-specific considerations such as passive stealth. We introduce selective, controllable sensor acquisition, where a user specifies a desired second sensor usage rate and the system acquires the additional sensor only when the primary sensor's estimated performance is sufficiently poor. An online threshold adjustment achieves the requested usage rate by adapting to recent primary sensor performance, directing additional sensing towards queries where fusion is most likely to help without calibration specific to sensor modality, PR method, or environment. The approach requires only a PR method that produces a match score and applies to both single-frame and sequence-based PR. Our system, across multiple datasets, PR descriptors, and sensor combinations, recovers 55\% of the available PR performance gain while using the second sensor on only 25\% of queries, and 81\% while using it on 50\%. We further analyse alternative acquisition decision mechanisms and usage-rate control methods, and examine performance when the second sensor is stronger than, comparable to, or weaker than the primary sensor.
\end{abstract}
\section{Introduction}
\label{sec:introduction}

Place recognition (PR) can benefit from both introspective methods that estimate whether a recognition result should be trusted, and multimodal approaches that combine information from multiple sensors~\cite{milford2026annurev,yin2026ToR,li2025air}. Introspective methods typically assess the performance of a primary sensor~\cite{carson2023icra,miller2026ral}, while multimodal PR improves robustness by fusing complementary sensor information~\cite{pan2026ral}. However, existing fusion approaches generally assume that observations from all required sensors have already been acquired and processed, with many learned architectures explicitly requiring paired or synchronised inputs~\cite{komorowski2021ijcnn,lai2022ral,wang2024prfusion}. Consequently, all sensors must be polled, encoded, and matched even when additional sensing contributes little to the final recognition result. This can be undesirable in deployed systems subject to constraints on energy, computation, or sensor utilisation~\cite{milford2026annurev,masone2021acce,ferrarini2019ahs,yin2026ToR}.

In this work, we instead treat multimodal PR as a \emph{selective acquisition} problem: information from a secondary sensor is acquired and processed only when the estimated performance of the primary sensor indicates that it is needed. We additionally allow the user to specify a desired \emph{duty}, defined as the fraction of queries for which the secondary sensor is used. Our method dynamically adapts the acquisition threshold from recent primary sensor performance estimates to achieve this requested duty. This concentrates additional sensing on queries for which fusion is most likely to be beneficial, without requiring separately calibrated thresholds for different sensor modalities, PR methods, or environments.

\begin{figure}[t]
\centering
\includegraphics[width=\columnwidth]{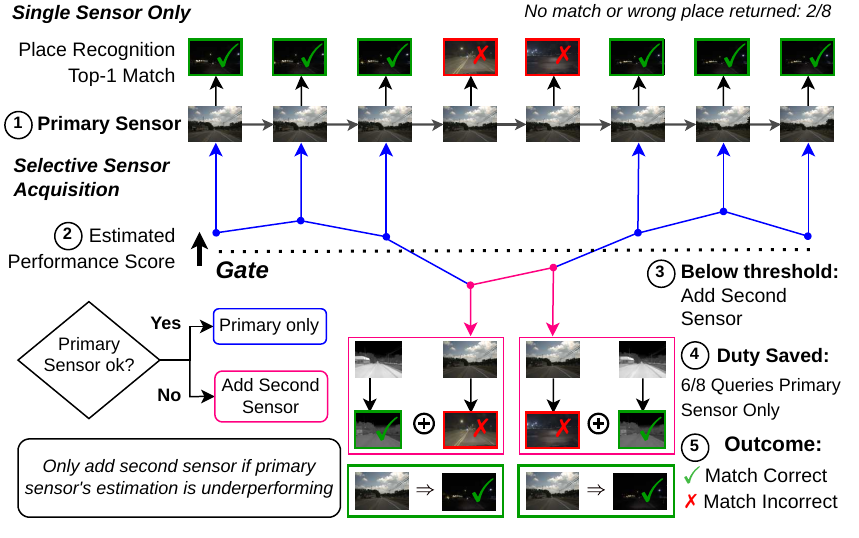}
\caption{\textbf{Selective Sensor Acquisition, query by query.} The primary sensor is processed (1), and the primary match estimated performance score is computed (2). Scores below a threshold trigger second sensor processing (3); only two of eight queries use that sensor (4). Adding the second sensor fixes the two selected primary matches (5).}
\label{fig:teaser}
\end{figure}

In particular, we make the following contributions:
\begin{itemize}
\item \textbf{Selective acquisition from primary sensor's estimated performance:}
We formulate multimodal PR as a selective acquisition problem and introduce a \emph{gate}, based on per-query estimated primary sensor performance that decides whether to process a second sensor observation \emph{before} it is polled, encoded or matched. 
We evaluate the gate with single-frame and sequence-based PR, comparing recognition accuracy against state-of-the-art acquisition methods~\cite{carson2023icra, yanko2026arxiv}, primary sensor introspection methods~\cite{zaffar2024cvpr}, and randomly selecting the second sensor at matched sensor use.
\item \textbf{A user-specified duty rate for a second sensor:}
We propose a recent-history-based method for dynamically adjusting primary sensor match acceptance threshold to achieve a desired overall utilisation duty for the second sensor.
\end{itemize}
The rest of the paper proceeds as follows. Sec.~\ref{sec:related-work} positions our approach within multimodal fusion, introspection, and conditional processing. Sec.~\ref{sec:methodology} defines the performance estimate, acquisition gate and duty control. Sec.~\ref{sec:experimentalsetup} describes the evaluation and experimental setup, Sec.~\ref{sec:results} presents performance, duty control and processing cost, and Sec.~\ref{sec:discussion} discusses our method, and proposes future work.
\section{Related Work}
\label{sec:related-work}
\looseness=-1
We position our work around three questions that guide our literature review: how place recognition systems use additional sensor information, how confidence estimates are used to decide what a system should do when a match is uncertain, and how we can constrain how often a second sensor is used.

\textbf{Using additional sensor information.}
Multimodal place recognition combines sensors with different failure modes to improve robustness as environmental conditions change~\cite{milford2026annurev,li2025air,yin2026ToR,carmichael2025ijrr, sferrazza2025cvprw}. Existing approaches primarily adapt how information from multiple sensors is combined, including learned or adaptive weighting~\cite{lai2022ral,pan2026ral}, late fusion of separately encoded modalities~\cite{komorowski2021ijcnn}, and attention-based or cross-modal feature interaction~\cite{zhou2023lcpr,wang2024prfusion,liang2025bdfusion,qi2025unimpr}. However, processing another sensor does not always improve the result~\cite{lai2022ral}. 

Sequential PR improves robustness by matching observations over time rather than treating each query independently~\cite{milford2012icra,garg2021corl}, and has also been extended to multimodal fusion~\cite{pan2026ral}. In all cases, the fusion decision is generally made \emph{after} the relevant sensor observations have already been acquired and processed. Our method decides whether to process the second sensor using only the primary sensor estimated performance, reducing computation and sensor usage when the additional sensing is not needed.

\looseness=-1
\textbf{Using estimated performance to decide what happens next.}
In robotic perception, introspection estimates the reliability of a system's own output~\cite{aldera2019itsc,yuan2023iros}, for example to guide navigation or provide map corrections. In recent place recognition works, confidence measures have similarly been used to predict whether a retrieved place match is correct~\cite{carson2022ral,miller2026ral,zaffar2024cvpr,carson2023icra}. These measures can be training-free~\cite{carson2022ral,miller2026ral} or learned from labelled data~\cite{cai2022iros,yanko2026arxiv}. Recent metrics use the highest-scoring reference matches~\cite{miller2026ral}, or measure the spread of a match's reference position~\cite{zaffar2024cvpr}. While low estimated performance is commonly used to reject or withhold an uncertain prediction~\cite{chow1970tit,claxton2024ral}, we instead use a per-query performance estimate to determine when additional information should be acquired from a second sensor.

\textbf{Working out or constraining how much a sensor is used.}
Conditional processing selects when to perform additional computation. Existing methods adjust image resolution~\cite{chaichi2026icassp}, select visual evidence using cached embeddings~\cite{hu2026arxiv}, or control fusion between sensors~\cite{lanegger2024exprob}. Online calibration addresses the related problem of adjusting thresholds to maintain a target prediction coverage as conditions change~\cite{gibbs2021neurips}. Our work instead adjusts a primary sensor's performance threshold to target a requested percentage of second sensor use.

\looseness=-1
Across these three areas, existing work either fuses sensors only after acquiring them~\cite{lai2022ral, pan2026ral}, uses confidence to withhold predictions~\cite{zaffar2021ijcv, miller2026ral,yanko2026arxiv}, or controls computation without targeting a rate the user has asked for~\cite{gibbs2021neurips}. Our method joins the three: a primary sensor performance estimate decides, before acquisition, whether the second sensor is processed, at a rate the operator specifies.
\section{Methodology}
\label{sec:methodology}
Our method operates on an existing PR pipeline. We first define the place recognition method, then describe how primary sensor performance is estimated for each query. We next use this estimate to decide whether to process the second sensor, before adjusting the acquisition threshold to target a user-defined percentage of second sensor use. Finally, we describe how the available primary and second sensor scores are combined to produce the final PR prediction.

\subsection{Place recognition}
\label{sec:meth:formulation}
We first define the PR system on which our acquisition method operates~\cite{schubert2023ram}. For $Q$ queries and $R$ reference places, let $\mathbf{P}\in\mathbb{R}^{R\times Q}$ denote the primary sensor matching score matrix, where $\mathbf{P}[r,q]$ is the similarity between query $q \in \{1,...,Q\}$ and reference place $r \in \{1,...,R\}$.

For single-frame place recognition, the primary sensor matching score vector for query $q$ is $\mathbf{d}_q=\mathbf{P}[:,q]$, where higher values indicate a better match.
For sequence-based recognition, the current scores are combined with those from the preceding $W$ queries:
\begin{equation}
\hat{\mathbf{d}}_q=(\mathrm{Seq}_W\mathbf{P})[:,q],
\qquad
(\mathrm{Seq}_W\mathbf{P})[r,q]
=
\sum_{\ell=0}^{W}
\mathbf{P}[r-\ell,q-\ell],
\label{eq:seq}
\end{equation}
where $W$ is the number of preceding queries included in the sequence and $\ell$ is the sequence offset, assuming that reference and query indices advance at the same rate~\cite{milford2012icra, garg2021corl}. 

Both $\mathbf{d}_q\in\mathbb{R}^{R}$ and $\hat{\mathbf{d}}_q\in\mathbb{R}^{R}$ are vectors of matching scores over the $R$ reference places, and the acquisition method treats them identically. From this point on, we write $\mathbf{d}_q$ for either. The two settings differ only in how an acquired second sensor observation is reused, which we return to in Sec.~\ref{sec:meth:fusion}.

\subsection{Primary sensor performance estimation}
\label{sec:meth:uncertainty}

Given the primary sensor matching scores $\mathbf{d}_q$, we compute a per-query \emph{performance estimate} $u_q$ that indicates how good the primary sensor's match is, relative to all the other places it could have matched. Here, larger values of $u_q$ indicate poorer estimated performance and therefore a greater need for second sensor information. The calculation uses only the primary sensor match and, where required, the known positions of the reference places. \emph{It does not require the second sensor query observation or the query's ground truth position.}

Our primary performance estimate is based on Spatial Uncertainty Estimation (SUE)~\cite{zaffar2024cvpr}, which asks whether the best-matching reference places agree on where the primary query match is. SUE takes the $k$ highest-scoring references in $\mathbf{d}_q$ and uses their known map positions. If those positions lie close together, the candidate match is likely correct. If they are spread across a trajectory, the sensor cannot distinguish places and therefore is considered less confident in its match.

Let $\mathbf{z}_{jq}$ be the known map position of the $j$th of those $k$ references, and let $\mathbf{\mu_q}$ be their weighted mean position. The SUE score is the total spread of the candidate positions about that mean:
\begin{equation}
u_q^{\mathrm{SUE}}
=
\operatorname{tr}(\boldsymbol{\Sigma}_q),
\qquad
\boldsymbol{\Sigma}_q
=
\frac{\sum_{j=1}^{k}w_{jq}(\mathbf{z}_{jq}-\boldsymbol{\mu}_q)(\mathbf{z}_{jq}-\boldsymbol{\mu}_q)^{\top}}{\sum_{j=1}^{k}w_{jq}},
\label{eq:sue}
\end{equation}
where $\operatorname{tr}(\cdot)$ is the matrix trace. Each candidate is weighted by how strongly it matches the query,
\begin{equation}
w_{jq}=\exp(-\lambda c_{jq}),
\label{eq:sue-weight}
\end{equation}
where $\lambda>0$ controls how quickly the weighting decreases and $c_{jq}$ is the cosine distance defined in Eq.~\ref{eq:cosine}.
\begin{equation}
c_{jq}
=
1-\frac{\mathbf{f}_q^{\top}\mathbf{f}_j}{\lVert\mathbf{f}_q\rVert\,\lVert\mathbf{f}_j\rVert}
\label{eq:cosine}
\end{equation}
is the cosine distance between the global descriptors $\mathbf{f}_q$ and $\mathbf{f}_j$ of query $q$ and reference candidate $j$. A large $u_q^{\mathrm{SUE}}$ means the candidate places are widely spread, so the primary sensor's match should not be trusted on its own.
We use the output of SUE as our performance estimate $u_q$, as one example of the introspective uncertainty scores available in the literature; Sec.~\ref{sec:res:baselines} compares it against several alternatives. In all cases the performance estimate is computed, and the acquisition decision made, \emph{before} the second sensor observation is polled, encoded or matched.

\subsection{Second sensor acquisition}
\label{sec:meth:gate}
As stated above, $u_q$ estimates primary sensor performance for query $q$, with larger values indicating poorer estimated performance. The acquisition gate compares this value with a threshold $\tau_q$:

\begin{equation}
m_q=\mathbf{1}[u_q>\tau_q]\in\{0,1\},
\label{eq:gate}
\end{equation}
where $\mathbf{1}[\cdot]$ is the indicator function. When $m_q=1$ the second sensor is encoded and matched; when $m_q=0$, it is not used for query $q$. A lower threshold acquires the second sensor more often, while a higher threshold acquires it less often.

The relationship between the numerical value of the performance estimate and whether a match should be trusted, and hence whether to go to the second sensor, varies by dataset, sensor combination and PR method. This makes setting a rigid acceptance threshold prior to deployment difficult. In the next section, we introduce a dynamic threshold adjustment process that deals with this challenge online.

\subsection{Requested duty control}
\label{sec:meth:policy}

We define the realised second sensor duty $D$ as the fraction of queries for which the second sensor is processed:
\begin{equation}
D
=
\frac{1}{Q}
\sum_{q=1}^{Q}
m_q.
\label{eq:duty}
\end{equation}
For example, $D=0.25$ means that the second sensor is processed for one quarter of the queries. The operator requests a target duty $D^\star\in(0,1)$, and the task of this section is to choose thresholds that deliver this target.

The performance estimates of a traverse form a vector $\{u_q\}_{q=1}^{Q}$, and where the threshold sits in that vector fixes the duty: raising it uses the second sensor less often, lowering it more often. We set the threshold in two ways.

\textbf{Offline}: if the whole traverse is available, we take the $(1-D^\star)$ quantile of all $Q$ scores:
\begin{equation}
\tau^{\mathrm{offline}}
=
\operatorname{quantile}_{1-D^\star}\!\left(\{u_q\}_{q=1}^{Q}\right),
\label{eq:tau-offline}
\end{equation}
which selects $\mathrm{round}(D^\star\cdot Q)$ queries, with seeded tie-breaking. This matches the requested duty to the nearest whole query, allowing state-of-the-art methods and our method to be compared at the same second sensor budgets in Sec.~\ref{sec:res:baselines}.

\textbf{Online}: if the scores are not known in advance, a fixed threshold does not give the same duty on every traverse: the spread of $u_q$ changes within a trajectory, as well as with datasets, sensor pairings, and PR descriptors. We therefore adapt the local receding window technique from~\cite{gibbs2021neurips}, which sets the threshold from recent scores and then corrects it using the decisions already made.

Let $\mathcal{H}_q$ be the preceding $L$ scores, $u_{\max(1,q-L)}$ to $u_{q-1}$. The current score is excluded, so the decision uses only information available before the current acquisition. We maintain a working acquisition level $\alpha_q$, initialised as $\alpha_1=D^\star$, and take the threshold as that quantile of the recent history:
\begin{equation}
\tau_q^{\mathrm{online}}=\operatorname{quantile}_{1-\alpha_q}(\mathcal{H}_q).
\label{eq:tauq}
\end{equation}
After each decision, the level is corrected by how far the gate sits from the request:
\begin{equation}
\alpha_{q+1}=\alpha_q+\gamma(D^\star-m_q),
\label{eq:duty-feedback}
\end{equation}
where $\gamma$ controls the strength of the correction. Using the second sensor lowers $\alpha_{q+1}$ and therefore raises the threshold. $L$ sets how quickly the threshold follows local changes in the score distribution, while $\gamma$ sets how quickly the realised duty is controlled.

This process targets an average acquisition duty rather than enforcing an exact number of acquisitions within every history window. The realised duty $D$ can therefore differ from the requested duty $D^\star$ when the uncertainty distribution changes, scores are tied, or the history is still being collected. We report both $D^\star$ and $D$ for the online setting in Sec.~\ref{sec:res:control}.

\subsection{Selective sensor score fusion}
\label{sec:meth:fusion}
Let $\mathbf{X}\in\mathbb{R}^{R\times Q}$ denote the second sensor matching score matrix that would be obtained if every second sensor query observation were processed. Writing $\mathbf{m}=(m_1,\ldots,m_Q)$ for the complete acquisition decision vector, only the second sensor score columns selected by the gate are available:
\begin{equation}
\mathbf{X}_{\mathbf{m}}[:,q]
=
\begin{cases}
\mathbf{X}[:,q], & m_q=1,\\
\mathbf{0}, & m_q=0,
\end{cases}
\label{eq:masked-scores}
\end{equation}
where $\mathbf{0}\in\mathbb{R}^{R}$ is an all-zero score vector, so a skipped query contributes nothing and $\mathrm{Seq}_W$ applies unchanged over sparse acquisitions.

We combine the available primary and second sensor scores with equal weight and obtain the final predicted reference place $\hat{\imath}_q$ for query $q$:
\begin{equation}
\hat{\imath}_q
=
\argmax_r
\left[
\mathrm{Seq}_W
\left(
\mathbf{P}
+
\mathbf{X}_{\mathbf{m}}
\right)
\right][r,q].
\label{eq:selective-sequence}
\end{equation}

For sequence-based recognition ($W>0$), an acquired second sensor score vector remains in the sequence window and can contribute to the following $W$ matches, whereas for single-frame recognition ($W=0$) it affects only the current query.

\section{Experimental Setup}
\label{sec:experimentalsetup}
We provide an overview of the datasets and PR methods used in our evaluation, followed by how we evaluate our method against the previous baselines, and how we parametrise our second sensor duty and how we determine our computational savings.
\subsection{Datasets and place recognition}
We evaluate on benchmark datasets that contain multiple traverses of the same route under varying conditions and, critically, time-aligned observations \emph{from more than one sensor modality}.
NSAVP~\cite{carmichael2025ijrr} covers two ${\sim}8$\,km routes with thermal, monochrome and, RGB cameras, under varied lighting and from opposing viewpoints.
ViViD++~\cite{lee2022ral} repeats the same spaces under changing illumination, with co-aligned RGB and thermal streams. MS$^2$~\cite{shin2023cvpr} covers city and suburban driving from morning to night with RGB, near-infrared (NIR) and, thermal imagery. Our main comparison uses the monochrome (mono), RGB, thermal and, NIR pairings.

For all datasets, sensor streams are temporally aligned and sampled at approximately $1$\,m intervals. Under the standard reference-query paradigm of place recognition research, we use day traverses as reference, and night traverses as the query test set~\cite{zaffar2021ijcv, schubert2023ram, carmichael2025ijrr}.

For every sensor pairing, we use ten global place recognition descriptors. Unless otherwise stated, all reported numbers represent the average performance of each of these approaches across all ten descriptors. The global place-recognition descriptors are: BoQ~\cite{Ali-bey_2024_CVPR},
CliqueMining~\cite{izquierdo2024closethereboostinggeographic},
CricaVPR~\cite{lu2024cvpr}, DINO-Mix~\cite{huangDINOMixEnhancingVisual2024},
EDTFormer~\cite{jin2025tcsvt}, MegaLoc~\cite{Berton_2025_MegaLoc},
MixVPR~\cite{alibey2023wacv}, QAA~\cite{xiao2026querybasedadaptiveaggregationmultidataset},
SALAD~\cite{Izquierdo_CVPR_2024_SALAD}, and SuperVLAD~\cite{lu2024supervlad},
using the VPR-methods-evaluation framework~\cite{berton2023vprmethods}.
We evaluate both single-frame ($W=0$) and sequence-based recognition, reporting $W=20$ as a representative sequence setting.

\subsection{Metrics}
We report Recall@1, where a query is correct when its top-1 retrieved reference lies within $10$\,m of its ground-truth position~\cite{zaffar2021ijcv,schubert2023ram}. Queries without a valid reference within this tolerance are excluded consistently across compared methods.

Second sensor use is reported as realised duty $D$. Acquisition methods are compared at matched duty: graphs report the full duty range, while tables report the specific operating points of $10\%$, $25\%$, and $50\%$ (Eq.~\eqref{eq:duty}). Recovery is calculated from dataset balanced mean recalls, with each of the three datasets weighted equally. 

\subsection{Baselines}
We have two broad types of comparator baselines. The first type of baseline is a randomised approach that is tuned to achieve the required duty cycle of the second sensor. Outperforming this baseline is not a claimed contribution: it is included primarily to provide a lower bound on performance.

The second type are the closest previous state-of-the-art techniques, which perform some form of introspective analysis, either through statistical factors~\cite{zaffar2024cvpr,carson2023icra} or through a learned model~\cite{yanko2026arxiv}. 

We include SUE~\cite{zaffar2024cvpr}, which we evaluate as a single sensor only baseline, following its original purpose: matches with low estimated performance are rejected, and no place prediction is returned for those queries. For the SUE primary sensor baseline, recall is calculated over all evaluated queries, with rejected queries counted as unsuccessful matches. We use the published weighting values from~\cite{zaffar2024cvpr} for non-sequential matching, and the sequence normalised scores for $W>0$. Our method instead uses additional second sensor information to attempt to resolve these queries.

Carson et al.~\cite{carson2023icra} estimate match quality from the shape of the similarity vector. It is training-free and descriptor-agnostic, but was developed to weight sequence matching. We adapt it to select queries for second sensor acquisition, sweeping its tolerance parameter to vary second sensor use.

KappaPlace-PT~\cite{yanko2026arxiv} is a learned comparator. It models descriptors as von Mises--Fisher variables and trains a lightweight head, using prototype-anchored supervision, to predict the concentration parameter as an estimate of aleatoric uncertainty. We use this estimate to guide second sensor acquisition. We tune sensor-combination weights on a separate dataset: MS$^2$ for evaluation on NSAVP and ViViD++, and NSAVP for evaluation on MS$^2$.

\begin{table*}[t]
\centering
\caption{Acquisition methods across second sensor duties and datasets using the offline duty sweep. Recall@1 (\%) is averaged across all sensor pairings and 10 global PR descriptors at $W=0$. For Single Sensor Only SUE, $D$ denotes the percentage of rejected queries, which are counted as misses. Means average the three dataset blocks. \textbf{Best} and \underline{second-best} scores are highlighted.
}
\label{tab:main}
\fontsize{8}{9.2}\selectfont
\setlength{\tabcolsep}{2pt}
\renewcommand{\arraystretch}{1.05}

\begin{tabular*}{\textwidth}{@{}l@{\extracolsep{\fill}}*{12}{r}@{}}
\toprule
& \multicolumn{4}{c}{10\% Duty} & \multicolumn{4}{c}{25\% Duty} & \multicolumn{4}{c}{50\% Duty} \\
\cmidrule(lr){2-5}\cmidrule(lr){6-9}\cmidrule(lr){10-13}
Method & NSAVP & MS$^2$ & ViViD++ & Mean & NSAVP & MS$^2$ & ViViD++ & Mean & NSAVP & MS$^2$ & ViViD++ & Mean \\
\midrule
Random & 82.84 & 75.22 & 82.80 & 80.28 & 84.21 & 76.37 & 85.42 & 82.00 & 86.51 & 78.34 & 89.75 & 84.87 \\
Primary Sensor Only -- SUE~\cite{zaffar2024cvpr} & 77.57 & 70.23 & 75.26 & 74.35 & 67.88 & 60.91 & 65.03 & 64.61 & 47.24 & 41.64 & 45.55 & 44.81 \\
Baseline -- Carson et al.~\cite{carson2023icra} & 82.97 & \underline{75.25} & 82.93 & \underline{80.39} & 84.54 & \underline{76.52} & 85.68 & \underline{82.25} & \underline{87.10} & \underline{78.61} & 90.10 & \underline{85.27} \\
Baseline -- KappaPlace-PT~\cite{yanko2026arxiv} & \underline{83.03} & 75.14 & \underline{82.98} & 80.38 & \underline{84.63} & 76.12 & \underline{85.85} & 82.20 & 87.07 & 78.01 & \underline{90.44} & 85.17 \\
Selective Acquisition (Ours) -- SUE & \textbf{85.11} & \textbf{77.13} & \textbf{84.70} & \textbf{82.32} & \textbf{87.91} & \textbf{79.19} & \textbf{89.19} & \textbf{85.43} & \textbf{89.99} & \textbf{80.85} & \textbf{94.39} & \textbf{88.41} \\
\bottomrule
\end{tabular*}
\end{table*}

\begin{figure*}[!t]
\centering
\includegraphics[width=\textwidth]{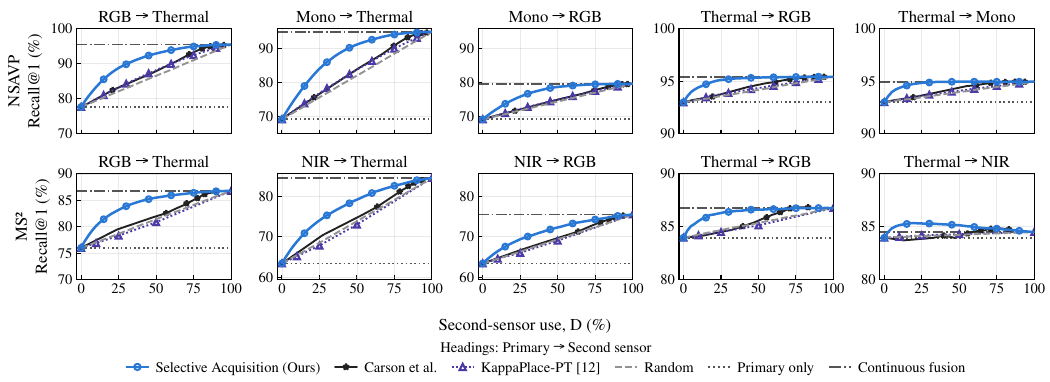}
\caption{Single-frame matching ($W=0$): Recall@1 versus second sensor use $D$ on NSAVP~\cite{carmichael2025ijrr} (top) and MS$^2$~\cite{shin2023cvpr} (bottom), averaged over ten PR descriptors in each panel. Columns correspond to sensor pairings, with the first sensor used as the primary sensor and the second as the duty-controlled second sensor. We compare our method with KappaPlace-PT~\cite{yanko2026arxiv}, Carson et al.~\cite{carson2023icra}, random acquisition, primary sensor only and continuous fusion.
}
\label{fig:main-signals-cells}
\end{figure*}

To assess how the choice of primary sensor estimated performance score affects selective acquisition, we compare several training-free introspection measures: Top match similarity~\cite{zaffar2024cvpr}, Lowe ratio~\cite{lowe2004ijcv}, Range margin~\cite{carson2022ral}, Peak sharpness~\cite{carson2023icra}, Perceptual Aliasing (PA) score~\cite{hausler2021arxiv}, and Statistical Uncertainty (SU)~\cite{miller2026ral}. We vary only the score used to guide acquisition, keeping the selection and fusion procedures fixed, and use the published weighting values where possible. This ablation tests which measures best identify queries that benefit from additional second sensor information at matched sensor use.

\subsection{Second sensor duty parameters}
For all online experiments, we use a local history of $L=400$ for the primary sensor queries and a feedback step of $\gamma=0.005$. The first ten queries are used to build the history before acquisition begins. We evaluate requested duties $D^{\star}$ of $10\%$, $25\%$, and $50\%$, and report the realised duty $D$ in each case. These settings are fixed across all datasets, sensor combinations, descriptors, and both single-frame and sequence-based recognition.

\subsection{Compute}
We measure per-query processing time on NVIDIA A100-SXM4-40GB with a batch size of one. Primary sensor only recognition provides the baseline cost in milliseconds. For queries that acquire the second sensor, we discuss the total processing time at matched requested second sensor duty.
\section{Results}
\label{sec:results}
\setcounter{dbltopnumber}{2}
\begin{figure*}[t]
\centering
\includegraphics[width=\textwidth]{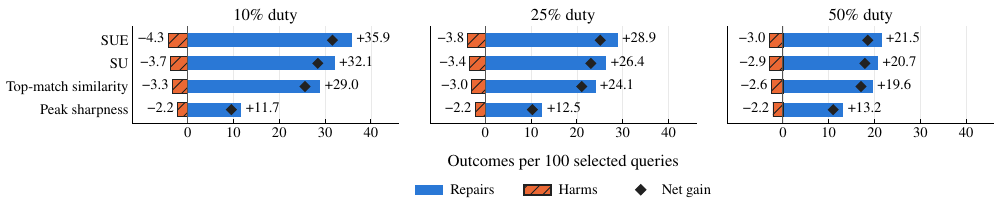}
\caption{Second sensor fusion outcomes per 100 selected queries at different duties. Diamonds show the net fusion gain: primary sensor errors corrected (blue bars) minus correct matches made incorrect (orange bars). Queries not represented within each 100 selected queries are unchanged after adding the second sensor. Compared metrics are SUE~\cite{zaffar2024cvpr}, Statistical Uncertainty (SU)~\cite{miller2026ral}, Top Match Similarity~\cite{zaffar2024cvpr}, and Peak Sharpness~\cite{carson2023icra}.
}
\label{fig:method-ratio}
\end{figure*}

\setcounter{topnumber}{1}
\AddToHook{shipout/after}[results-tables]{%
  \ifnum\value{page}=\getpagerefnumber{tab:baselines}\relax
    \setcounter{topnumber}{2}%
    \RemoveFromHook{shipout/after}[results-tables]%
  \fi
}
We extensively evaluate our method across 20 sensor pairings, three datasets, and ten global PR descriptors, comparing recall improvements against multiple state-of-the-art methods and random acquisition baselines. We also present the ability of the system to dynamically achieve a desired duty cycle, a comparison to an upper bound offline approach, and evaluate the limitations of how poor the second sensor must be before the selectively polled and fused performance of our performance degrades.

\subsection{Performance gain at different duty cycles}
\label{sec:res:baselines}
At the same acquisition budgets, our method achieves higher mean Recall@1 than random acquisition and the compared baselines (Table~\ref{tab:main}). For example, at $10\%$ duty it raises Recall@1 from $79.16\%$ with the primary sensor alone to $82.32\%$, against $80.28\%$ for random acquisition, $80.39\%$ for Carson~\cite{carson2023icra} and $80.38\%$ for the trained KappaPlace-PT~\cite{yanko2026arxiv}. Our method also achieves the highest mean Recall@1 at the $25\%$ and $50\%$ duties reported in the table. 

Our method consistently outperforms the state-of-the-art and random baselines across the full duty sweep, with the magnitude of the performance gain ultimately determined by the second sensor duty. Figure~\ref{fig:main-signals-cells} shows results for individual NSAVP and MS$^2$ sensor pairings, averaged over ten PR descriptors. Each curve is a sweep of the acquisition threshold, covering second sensor use from 0\% single sensor only to 100\% continuous fusion.

We repeated the same experiments, only changing the sequential window as specified in Sec.~\ref{sec:meth:fusion}, to $W=20$. In this sequential scheme, our method achieves higher mean Recall@1 than random acquisition and all compared baselines at the same acquisition budgets (Table~\ref{tab:w20-method-means}). For example, at $10\%$ duty, it raises Recall@1 from $88.50\%$ with the primary sensor alone to $91.57\%$, against $90.80\%$ for random acquisition, $90.95\%$ for Carson-based
acquisition~\cite{carson2023icra} and $90.29\%$ for the trained KappaPlace-PT~\cite{yanko2026arxiv}. At $50\%$ duty, our method reaches $93.28\%$, recovering approximately $96\%$ of the achievable PR performance gain.

We next ask which primary sensor performance estimate best identifies queries where processing the second sensor improves recognition. We compare the outcomes of acquisition selected by alternative metrics that produce a performance score. Figure~\ref{fig:method-ratio} reports, per 100 selected queries, how many primary sensor errors fusion corrects and how many correct primary sensor matches fusion makes incorrect.

SUE produces the largest net gain across the evaluated duties and achieves the highest Recall@1 in Table~\ref{tab:baselines}.

SUE~\cite{zaffar2024cvpr} and SU~\cite{miller2026ral} select more useful acquisitions than the ratio- and margin-based measures~\cite{lowe2004ijcv,carson2022ral,carson2023icra}, which provide only small improvements over random acquisition. These results motivate our use of SUE as the primary sensor performance estimate in the remaining experiments.

\begin{table}[t]
\centering
\caption{Acquisition methods across second sensor use duties and datasets run as an offline duty sweep at $W=20$. Mean Recall@1 (\%), with same parameters as Table~\ref{tab:main}.}
\label{tab:w20-method-means}
\fontsize{8}{9.2}\selectfont
\setlength{\tabcolsep}{2pt}
\renewcommand{\arraystretch}{1.05}

\begin{tabular*}{\columnwidth}{@{}>{\raggedright\arraybackslash}p{.59\columnwidth}@{\extracolsep{\fill}}rrr@{}}
\toprule
& \multicolumn{3}{c}{Duty $D$ (\%)} \\
\cmidrule(lr){2-4}
Method & 10 & 25 & 50 \\
\midrule
Random & 90.80 & \underline{92.30} & 93.05 \\
Primary Sensor Only -- SUE~\cite{zaffar2024cvpr} & 81.64 & 69.20 & 46.80 \\
Baseline -- Carson et al.~\cite{carson2023icra} & \underline{90.95} & \underline{92.30} & \underline{93.09} \\
Baseline -- KappaPlace-PT~\cite{yanko2026arxiv} & 90.29 & 91.55 & 92.67 \\
Selective Acquisition (Ours) -- SUE & \textbf{91.57} & \textbf{92.67} & \textbf{93.28} \\
\bottomrule
\end{tabular*}
\end{table}

\begin{table}[t]
\centering
\caption{Average Recall@1 (\%) at matched second sensor offline duty $D$ for $W=0$ over 10 PR global descriptors and all sensors and datasets. \textbf{Best} and \underline{second} rank confidence scores.}
\label{tab:baselines}
\fontsize{8}{9.2}\selectfont
\setlength{\tabcolsep}{2pt}
\renewcommand{\arraystretch}{1.05}

\begin{tabular*}{\columnwidth}{@{}>{\raggedright\arraybackslash}p{.59\columnwidth}@{\extracolsep{\fill}}rrr@{}}
\toprule
 & \multicolumn{3}{c}{Duty $D$ (\%)} \\
\cmidrule(lr){2-4}
Performance Estimate Method & 10 & 25 & 50 \\
\midrule
Random & 80.28 & 82.00 & 84.87 \\
SUE~\cite{zaffar2024cvpr} & \textbf{82.32} & \textbf{85.43} & \textbf{88.41} \\
SU~\cite{miller2026ral} & \underline{82.00} & \underline{84.91} & \underline{88.10} \\
PA~\cite{hausler2021arxiv} & 80.68 & 82.84 & 86.02 \\
Top Match Similarity~\cite{zaffar2024cvpr} & 81.72 & 84.42 & 87.68 \\
Lowe ratio~\cite{lowe2004ijcv} & 80.42 & 82.29 & 85.24 \\
Range margin~\cite{carson2022ral} & 80.46 & 82.35 & 85.36 \\
Peak sharpness~\cite{carson2023icra} & 80.11 & 81.73 & 84.65 \\
\bottomrule
\end{tabular*}
\end{table}

\subsection{A user-defined second sensor duty rate can control how often the second sensor is used}
\label{sec:res:control}
We previously set $\tau_q$ using an offline duty sweep, as a fair comparison with the previous state-of-the-art offline benchmarks. This shows how the system performs across the duty range, but it does not let a user choose in advance where in the range to operate. 

\begin{figure}[t]
\centering
\includegraphics[width=\columnwidth]{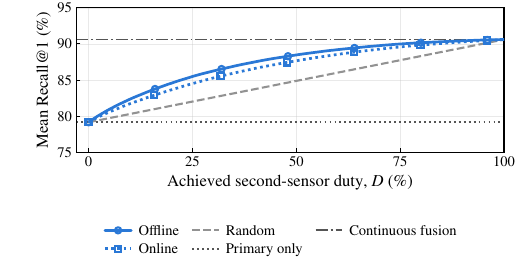}
\caption{Mean Recall@1 versus achieved second sensor duty for offline duty specified sensor acquisition and online receding-window acquisition. Results are the average of three datasets, 20 sensor pairs, and 10 PR global descriptors. Baselines show random acquisition, primary sensor only matching, and continuous fusion.}
\label{fig:offline-online}
\end{figure}

Instead, using the receding window threshold adjustment of Sec.~\ref{sec:meth:policy}, we show that a requested duty $D^\star$ can be achieved using a local window, controlling the rate at which the threshold is raised or lowered according to local conditions along a traverse. Across all sensor configurations, requests of $D^\star=10\%$, $25\%$ and $50\%$ achieved realised duties $D$ within a $\pm1\%$ tolerance, and Recall@1 online is only marginally below the offline result at the same duty (Figure~\ref{fig:offline-online}). This online adjustment is also insensitive to the window length $L$: Recall@1 varied by less than 1\%, showing that only a short history is required to control second sensor use effectively.

\begin{figure}[t]
\centering
\includegraphics[width=\columnwidth]{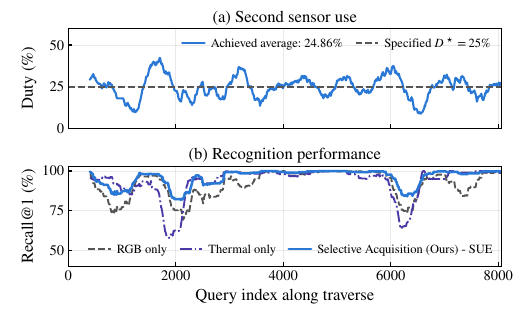}
\caption{Online second sensor use on NSAVP~\cite{carmichael2025ijrr} with SALAD~\cite{Izquierdo_CVPR_2024_SALAD}, RGB primary and thermal optional second sensor. Top: sliding-window duty cycle for a requested $D^\star=25\%$, achieving $24.86\%$ over the traverse. Bottom: Recall@1 for RGB only, thermal only and our selective acquisition method. Both panels use trailing $400$-query windows.}
\label{fig:duty-control}
\end{figure}

The effect of the local window adjusting along a traverse is best illustrated in Figure~\ref{fig:duty-control}. Although acquisition varies locally along a traverse, the system adjusts the threshold in regions where sensor performance degrades, increasing second sensor use when needed. In regions where both individual sensors perform poorly, our method can still achieve higher Recall@1 than either sensor alone.

\begin{figure}[t]
\centering
\includegraphics[width=\columnwidth]{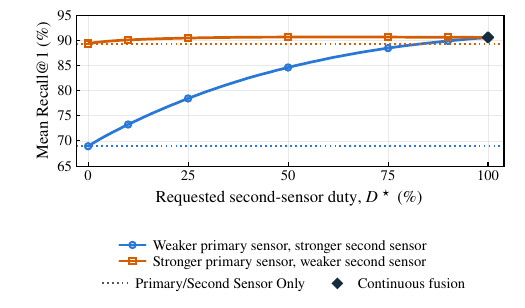}
\caption{Effect of sensor roles on online SUE acquisition ($W=0$). Stronger and weaker sensors are identified by their Recall@1. This cohort includes all sensor pairings, across the 10 PR descriptors. Dotted lines show recall and the diamond shows continuous fusion.}
\label{fig:A-B-B-A}
\end{figure}

\subsection{Less processing time and fewer sensor calls at matched recall}
\label{sec:res:cost}
We compare computational cost by matching each baseline to the Recall@1 achieved by our method at $25\%$ requested second sensor use during an online controller. Averaged over the sensor and PR descriptor methods, this corresponds to $83.34\%$ mean Recall@1. At this matched performance, our method requires an average of $29.40$\,ms per query on an A100 GPU, compared with $33.25$\,ms for random acquisition, $31.96$\,ms for Carson et al.~\cite{carson2023icra}, and $32.74$\,ms for KappaPlace-PT~\cite{yanko2026arxiv}, corresponding to an $8$--$12\%$ reduction in average processing time.

To achieve this performance, our method processes the second sensor on only $24.75\%$ of queries, compared with $50.67\%$ for random acquisition, $43.61\%$ for Carson, and $48.22\%$ for KappaPlace-PT. Thus, the baselines require $76$--$105\%$ more second sensor calls than our method to achieve the same Recall@1.

\subsection{Understanding when selection helps}
\label{sec:res:limits}
In Figure~\ref{fig:A-B-B-A}, we examine how the relative standalone PR performance of the two sensors affects the benefit of selective acquisition. 
When the primary sensor's recall is weak and the second sensor is strong, selective acquisition substantially improves mean Recall@1, from 68.9\% to 73.3\% at just 10\% requested duty and 78.5\% at 25\% duty, reaching 90.6\% with continuous fusion.
When the primary sensor's recall is strong and the second sensor is weaker, selective acquisition still improves performance but offers relatively limited gains: mean Recall@1 increases from 89.4\% to 90.5\% at 25\% requested second sensor duty, then plateaus near 90.6\%.

Our system estimates primary sensor confidence rather than the expected benefit of the second sensor, since that observation has not yet been processed. It can therefore determine \emph{when} additional information is likely to be useful, including during local degradation of an otherwise high-performing primary sensor, but not whether a particular second sensor will provide useful complementary evidence. 

We leave assessing the suitability of the second sensor's contribution as future work, as doing so would require the second sensor to be polled and processed before its usefulness could be determined.

\section{Discussion and Conclusion}
\label{sec:discussion}
Place recognition systems conventionally use all the information available to decide whether a match should be trusted or rejected. We instead use the primary sensor's estimated performance to decide when additional sensor information is worth acquiring at all. In addition, because the relationship between that estimate and a trustworthy match is inconsistent across PR methods, sensors and, environmental conditions, a threshold fixed ahead of deployment does not transfer. We therefore let the operator specify how often the second sensor should be used, and derive the acquisition threshold from that request. This duty based formulation is agnostic to the sensor's modality, the dataset, and the performance of the primary sensor.

We demonstrate that our system recovers $55\%$ and $81\%$ of the Recall@1 gain available from continuous fusion while using the second sensor on $25\%$ and $50\%$ of queries, respectively. At matched duty, our method achieves higher mean Recall@1 than the state-of-the-art introspection baselines, 82.32\% at 10\% duty against 80.39\% for Carson et al.~\cite{carson2023icra} and 80.38\% for the trained KappaPlace-PT~\cite{yanko2026arxiv}. 

No previous system lets the operator specify a sensor duty in advance, whereas our online controller reaches a requested duty within $\pm1\%$ across all sensor configurations, at a small cost in Recall@1 relative to the same duty selected offline. 

Importantly, the primary sensor need not be the sensor with the highest Recall@1. In deployment, a lower-performing sensor may be preferred for continuous operation because of power, computational, update-rate, or application-specific constraints, while a higher-performing but more costly second sensor is acquired selectively according to the user-defined requested duty.

The key benefit of the primary sensor performance estimate is that the system requires neither ground truth nor the second sensor observation to decide whether additional information should be acquired. The user can specify how often the second sensor is used in advance, allowing the cost of second sensor acquisition to be controlled before that acquisition cost is incurred.

Future work could extend selective acquisition to multiple optional sensors and jointly consider expected recognition benefit, energy, latency, and acquisition cost when deciding which additional information to acquire.

{\footnotesize
\let\oldthebibliography\thebibliography
\renewcommand{\thebibliography}[1]{%
  \oldthebibliography{#1}%
  \setlength{\itemsep}{0pt plus 0.2pt}%
  \setlength{\parsep}{0pt}}
\FloatBarrier
\bibliographystyle{IEEEtran}
\bibliography{ExportedItems}
}

\end{document}